\documentclass[final,5p,times,twocolumn]{elsarticle}

\usepackage[utf8]{inputenc} 
\usepackage[T1]{fontenc}    
\usepackage{hyperref}       
\usepackage{url}            
\usepackage{booktabs}       
\usepackage{amsfonts}       
\usepackage{nicefrac}       
\usepackage{microtype}      
\usepackage{amsmath,amssymb} 
\usepackage{bbold}
\usepackage{color}
\usepackage{wrapfig}
\usepackage{pgffor}
\usepackage{tabularx}
\usepackage{float}
\usepackage{stfloats}

\makeatletter
\def\ps@pprintTitle{%
    \let\@oddhead\@empty
    \let\@evenhead\@empty
    \def\@oddfoot{\myfooter}
    \let\@evenfoot\@oddfoot
}
\makeatother

\def\myfooter{
  {\footnotesize
  \begin{minipage}{\textwidth}
  Pattern Recognition 78 (2018) 331–343, \url{https://doi.org/10.1016/j.patcog.2018.01.036}.\\
  \copyright~2018. This manuscript version is made available under the CC-BY-NC-ND 4.0 license http://creativecommons.org/licenses/by-nc-nd/4.0/
  \end{minipage}
  }
}

\def\X{{\bf X}}
\def\x{{\bf x}}

\def\C{{\bf C}}
\def\V{{\bf V}}
\def\S{{\bf S}}

\def\W{{\bf W}}

\def\d{{\bf d}}

\def\bigO2{\mbox{${\cal O}$}}
\def\bigO{O}

\def\1n{\mathbf{1}_n}
\def\0{\mathbf{0}}
\def\1{\mathbf{1}}

\def\C{{\bf C}}

\def\S{{\bf S}}

\def\V{{\bf V}}
\def\W{{\bf W}}
\def\X{{\bf X}}

\def\n{{\bf n}}

\def\a1{\mbox{\bf a}_1}
\def\a2{\mbox{\bf a}_2}
\def\a3{\mbox{\bf a}_3}
\def\a4{\mbox{\bf a}_4}

\DeclareMathOperator*{\argmin}{\arg\!\min}

\def\vec{\mathop{\rm vec}}								
\def\tr{\mathop{\rm tr}}								
\def\diag{\mathop{\rm diag}}								

\journal{Pattern Recognition}

\begin{document}

\begin{frontmatter}



\title{Deep Unsupervised Learning of Visual Similarities}


\author[]{Artsiom Sanakoyeu\corref{cor1}}
\cortext[cor1]{Corresponding author.}
\ead{artsiom.sanakoyeu@iwr.uni-heidelberg.de}
\author{Miguel A. Bautista}
\ead{miguel.bautista@iwr.uni-heidelberg.de}
\author{Bj{\"o}rn Ommer}
\ead{bjoern.ommer@iwr.uni-heidelberg.de}
\address{Heidelberg Collaboratory for Image Processing and Interdisciplinary Center for Scientific Computing, Heidelberg University, Germany}

\begin{abstract}
Exemplar learning of visual similarities in an unsupervised manner is a problem of paramount importance to Computer Vision.
In this context, however, the recent breakthrough in deep learning could not yet unfold its full potential. 
With only a single positive sample, a great imbalance between one positive and many negatives, and unreliable relationships between most samples, training of Convolutional Neural networks is impaired. 
In this paper we use weak estimates of local similarities and propose a single optimization problem to extract batches of samples with mutually consistent relations.
Conflicting relations are distributed over different batches and similar samples are grouped into compact groups.
Learning visual similarities is then framed as a sequence of categorization tasks.
The CNN then consolidates transitivity relations within and between groups and learns a single representation for all samples without the need for labels.
The proposed unsupervised approach has shown competitive performance on detailed posture analysis and object classification.

\end{abstract}


\end{frontmatter}


\section{Introduction}

Learning similarities in the visual domain plays a central role for numerous computer vision tasks which range across different levels of abstraction, from low-level image processing to high-level object recognition or human pose estimation.
Similarities have been usually obtained as a result of 
category-level recognition, where categories and the similarities of all their samples to other classes are jointly modeled.
However, the large intra-class variability of visual categories has recently spurred exemplar methods \cite{exemplarsvm,exemplarsvm2}, which split the category-level model into simpler sub-tasks for each sample.
Therefore, separate exemplar classifiers are trained by learning the similarities of individual exemplars against a large set of negatives.
This paradigm of exemplar learning has been applied with successful results in problems like object recognition \cite{exemplarsvm,angelacvpr14}, instance retrieval \cite{paris,josepr}, and grouping \cite{hoglda}.
Learning visual similarities has been also of particular importance for posture analysis \cite{posesearch} and video parsing \cite{videoparsing}, where exploiting both the appearance \cite{ConvNetpretext1} and the temporal domain \cite{ConvNetpretext2} has proven useful.
 
Throughout the numerous methods for learning visual similarities, supervised techniques have been of particular interest in the computer vision field.
These supervised techniques have therefore followed different formulations either as ranking \cite{simlearning1}, regression \cite{simregression}, and classification \cite{videoparsing} problems.
Furthermore, with the recent advent of Convolutional Neural Networks (CNN), two stream architectures \cite{ConvNetSimPatch} and ranking losses \cite{ConvNetSimTriplet} have shown great improvements over similarities learned using hand-crafted features.
Nevertheless, these performance improvements obtained by CNNs come at the cost of requiring millions of samples of supervised training data or at least the fine-tuning \cite{ConvNetpretext1} on large labeled datasets such as PASCAL VOC.
Even though the amount of accessible image data is growing at an ever increasing rate, supervised labeling of image similarities is extremely costly.
In addition to the difficulty of labeling a similarity metric, not only similarities between images are important, but also between objects and their parts.
Annotating the fine-grained similarities between all these entities is hopelessly complex, in particular for the large datasets typically used for training CNNs.

Unsupervised deep learning of similarities that does not require any labels for pre-training or fine-tuning is, therefore, of great interest to the vision community.
This way we can utilize large image datasets without being limited by the need for costly manual annotations.
However, CNNs for exemplar-based learning have been rare \cite{exemplarcnn} due to limitations resulting from the widely used cross-entropy loss.
The learning task suffers from only a single positive instance, it is highly unbalanced with many more negatives, and the relationships between samples are unknown, cf.
Sec.~\ref{sec:methodology}.
Consequentially, stochastic gradient descend (SGD) gets corrupted and has a bias towards negatives, thus forfeiting the benefits of deep learning.

Our approach overcomes this shortcoming by updating similarities and CNN parameters.
Normally, at the beginning, only a few local estimates of similarities are easily available (i.e. pairs of samples that are highly similar (near duplicates) or that are very distant).
Nevertheless, most of the initial similarities are unknown, or non-transitive, i.e. mutually contradicting.
To nevertheless define balanced classification tasks suited for CNN training, 
we formulate an optimization problem that builds training batches for the CNN by selecting groups of compact cliques so that all cliques in a batch are mutually distant.
Thus for all samples of a batch (dis-)similarity is defined---they either belong to the same 
compact clique or are far away and belong to different cliques.
However, pairs of samples with no reliable similarities end up in different batches so they do not yield false training signal for SGD.
Classifying if a sample belongs to a clique serves as a pretext task for learning exemplar similarity.
Training the network then implicitly reconciles the transitivity relations between samples in different batches.
Thus, the learned CNN representations impute similarities that were initially unavailable and generalize them to unseen data.
Furthermore, to incorporate temporal context in our model, 
we introduce a Local Temporal Pooling strategy that models how similarities between exemplars change over short periods of time.

In the experimental evaluation, the proposed approach significantly improves over state-of-the-art approaches for posture analysis and retrieval by learning a general feature representation for a human pose that can be transferred across datasets.

\subsection{Related Work} \label{sec:relatedwork}

The Exemplar Support Vector Machine (Exemplar-SVM) has been one of the driving methods for exemplar-based learning \cite{exemplarsvm}.
Each Exemplar-SVM classifier is defined by a single positive instance and a large set of negatives.
To improve performance, Exemplar-SVMs require several rounds of hard negative mining, increasing greatly the computational cost of this approach.
To circumvent this high computational cost, \cite{hoglda} proposes to train Linear Discriminant Analysis (LDA) over Histogram of Gradient (HOG) features \cite{hoglda}.
LDA whitened HOG features with the common covariance matrix estimated for all the exemplars removes correlations between the HOG features, which tend to amplify the background of the image.

Recently, several CNN approaches have been proposed for supervised similarity learning using either pairs \cite{ConvNetSimPatch}, or triplets \cite{ConvNetSimTriplet} of images.
However, supervised formulations for learning similarities require that the supervisory information scales quadratically for pairs of images, or cubically for triplets.
This results in very large training times. 

The literature on exemplar-based learning in CNNs is very scarce.
In \cite{exemplarcnn} the authors of Exemplar-CNN tackle the problem of unsupervised feature learning.
A patch-based categorization problem is designed by randomly extracting patch for each image in the training set and defining it as a surrogate class.
Hence, since this approach does not take into account (dis-)similarities between exemplars, it fails to model their transitivity relationships, resulting in poor performances (see Sect. \ref{sec:olympic_metrics}).

Furthermore, recent works \cite{ConvNetpretext2}, \cite{ConvNetpretext1}, \cite{hyvarinenICA} and \cite{shuffleandlearn}
showed that temporal information in videos and spatial context information in images can be utilized as a convenient supervisory signal for learning feature representation with CNNs.
However, the computational cost of the training algorithm is enormous since the approach in \cite{ConvNetpretext1} 
needs to tackle all possible pair-wise image relationships requiring a 
training set that scales quadratically with the number of samples.
On \cite{hyvarinenICA} authors leverage time-contrastive loss to learn representations leveraging the temporal structure of the data.
However, this approach is limited to video sequences without repetitions since the method is based on the assumption of mutual independence of time segments.
In contrast, our approach leverages the relationship information between compact cliques, framing similarity learning as a multi-class classification problem.
As each training batch contains mutually distinct cliques the computational cost of the training algorithm is greatly decreased.

\section{Methodology} \label{sec:methodology}

In this section we show how a CNN can be employed for learning similarities between all pairs of a large number of exemplars.
In particular, the idiosyncrasies of exemplar learning have made it difficult to unravel its full capabilities in CNNs.
First, deep learning is extremely data hungry, which conflicts with having a single positive exemplar for training, we now abbreviate this setup as 1-sample CNN.
This 1-sample setup then faces several issues.
\emph{(i)} The within-class variance of an individual exemplar cannot be modeled.
\emph{(ii)} The ratio of one exemplar and many negatives is highly imbalanced so that the cross-entropy loss over SGD batches overfits against the negatives.
\emph{(iii)} An SGD batch for training a CNN on multiple exemplars can contain arbitrarily similar samples with different label (the different exemplars may be similar or dissimilar), resulting in label inconsistencies.
\emph{(iv)} Provided the single training sample, exemplar learning cannot exploit the temporal context of training data, if available.

The methodology proposed in this paper overcomes this issues as follows.
In Sect.~\ref{sec:cliqueFormation} we discuss why simply appending an exemplar with its nearest neighbors and data augmentation (similar in spirit to the Clustered Exemplar-SVM \cite{clusteredsvm}, which we abbreviate as NN-CNN) is not sufficient to address \emph{(i)}.
 Sect.~\ref{sec:batchselection} deals with \emph{(ii)} and \emph{(iii)} by generating batches of cliques that maximize the intra-clique similarity while minimizing inter-clique similarity.
In addition, Sect. \ref{sec:TemporalContext} shows how to exploit temporal information to further impose structure on the learned similarities by using a temporal average pooling.

\begin{figure*}[t!]
    \centering
    \setlength{\tabcolsep}{0 pt}
    \begin{tabular}{cccc}
    \includegraphics[width=0.25\textwidth, height = 3 cm]{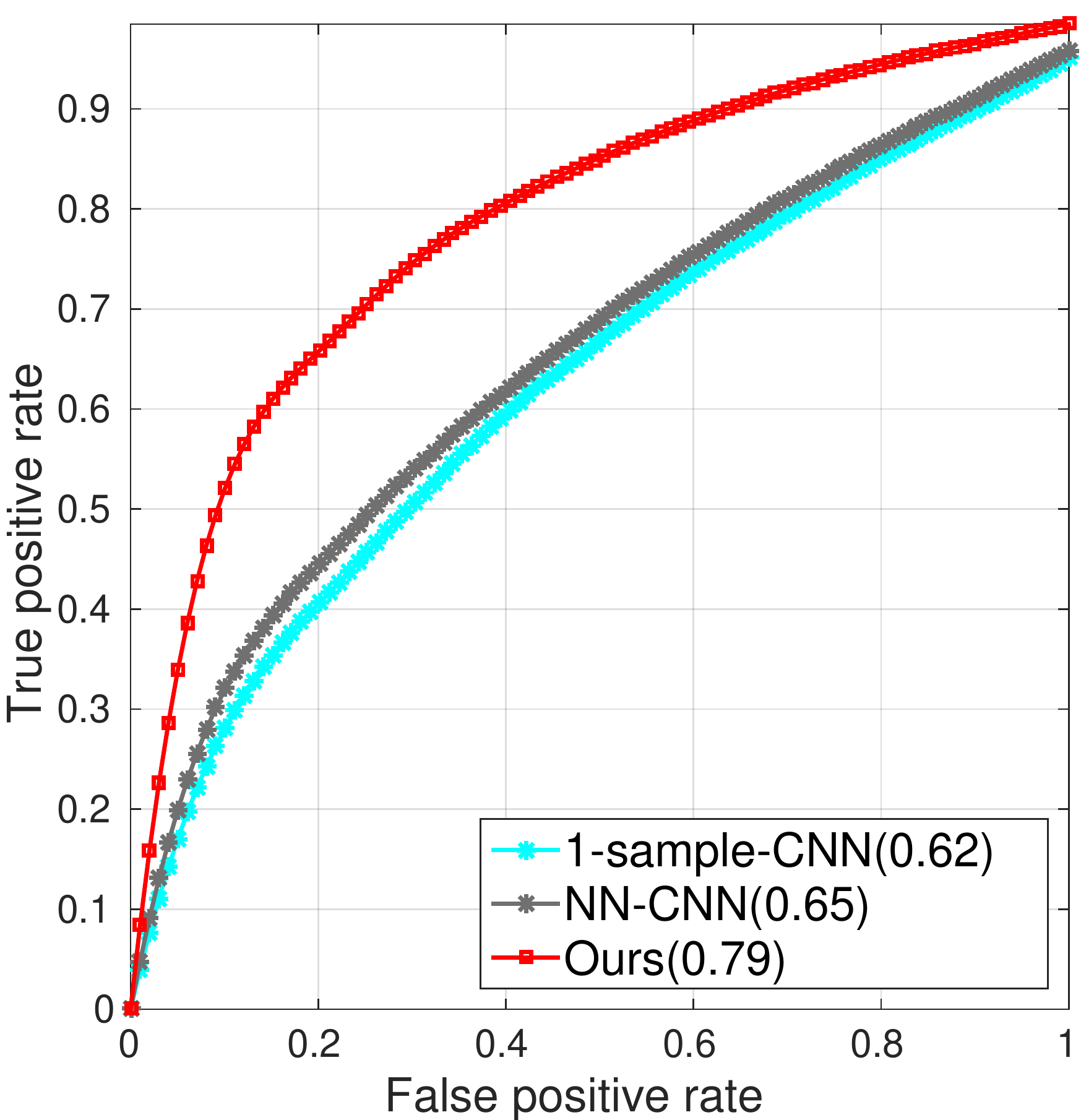} &  \includegraphics[width=0.25\textwidth, height = 3 cm]{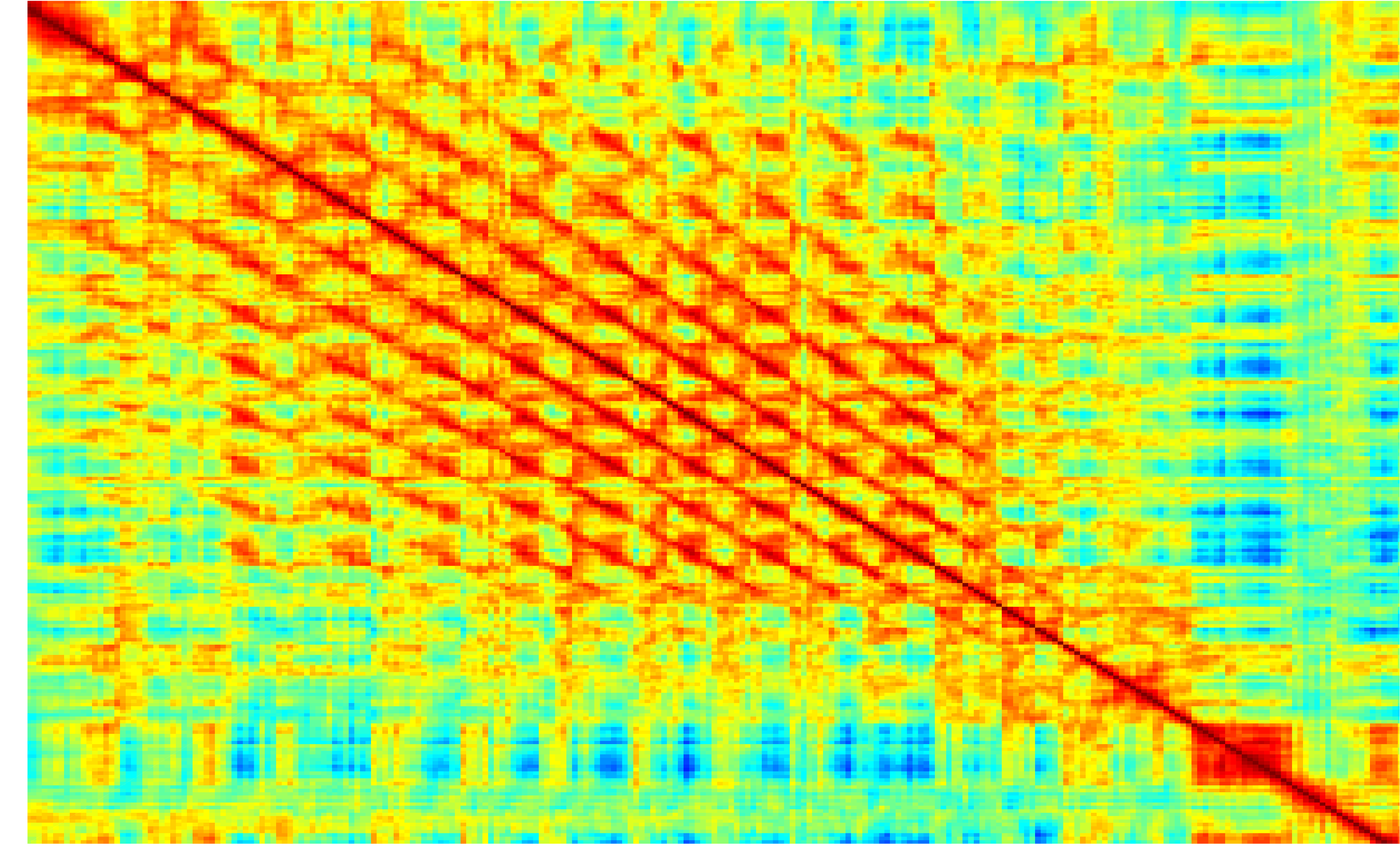} &  \includegraphics[width=0.25\textwidth, height = 3 cm]{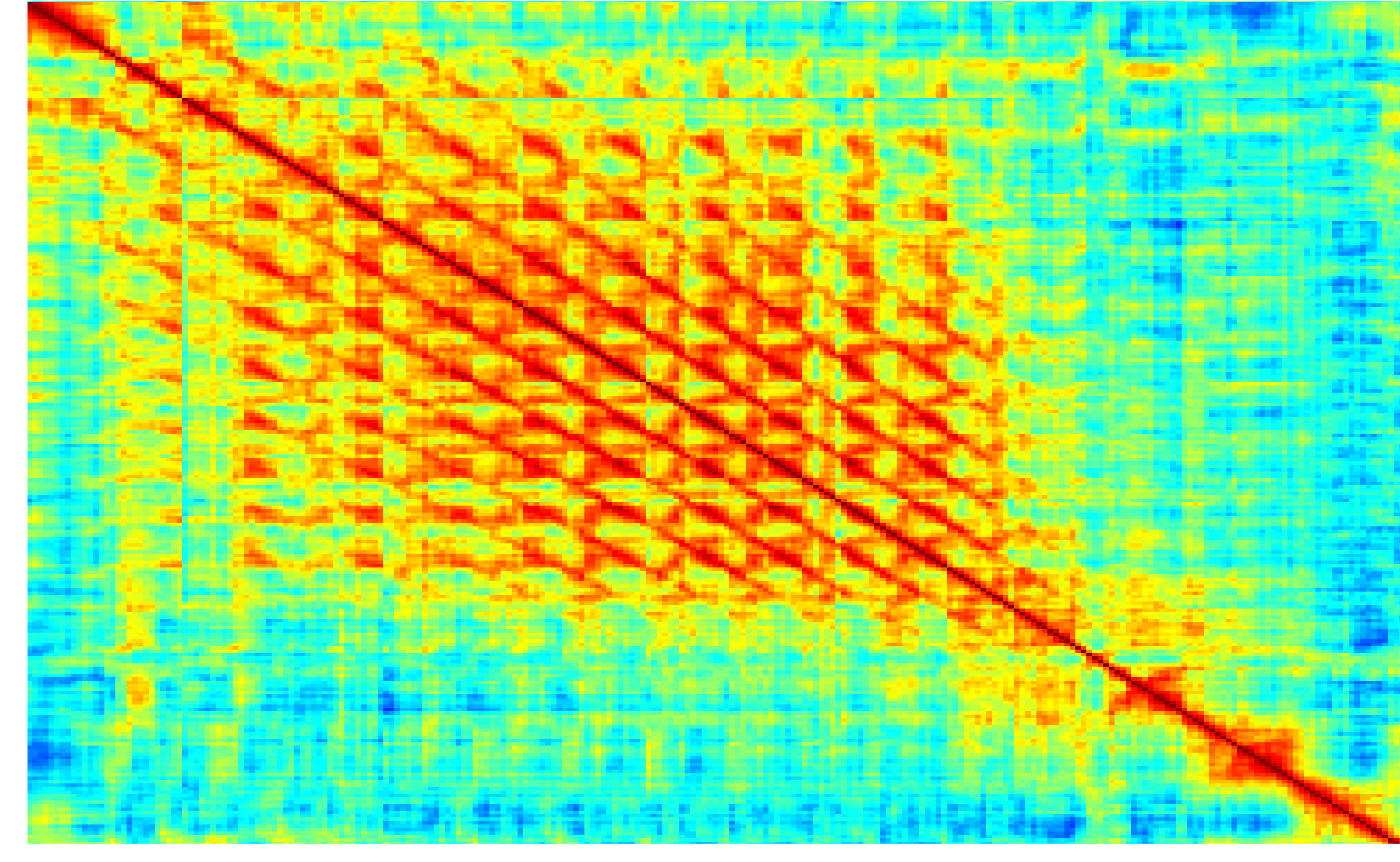} &  \includegraphics[width=0.25\textwidth, height = 3 cm]{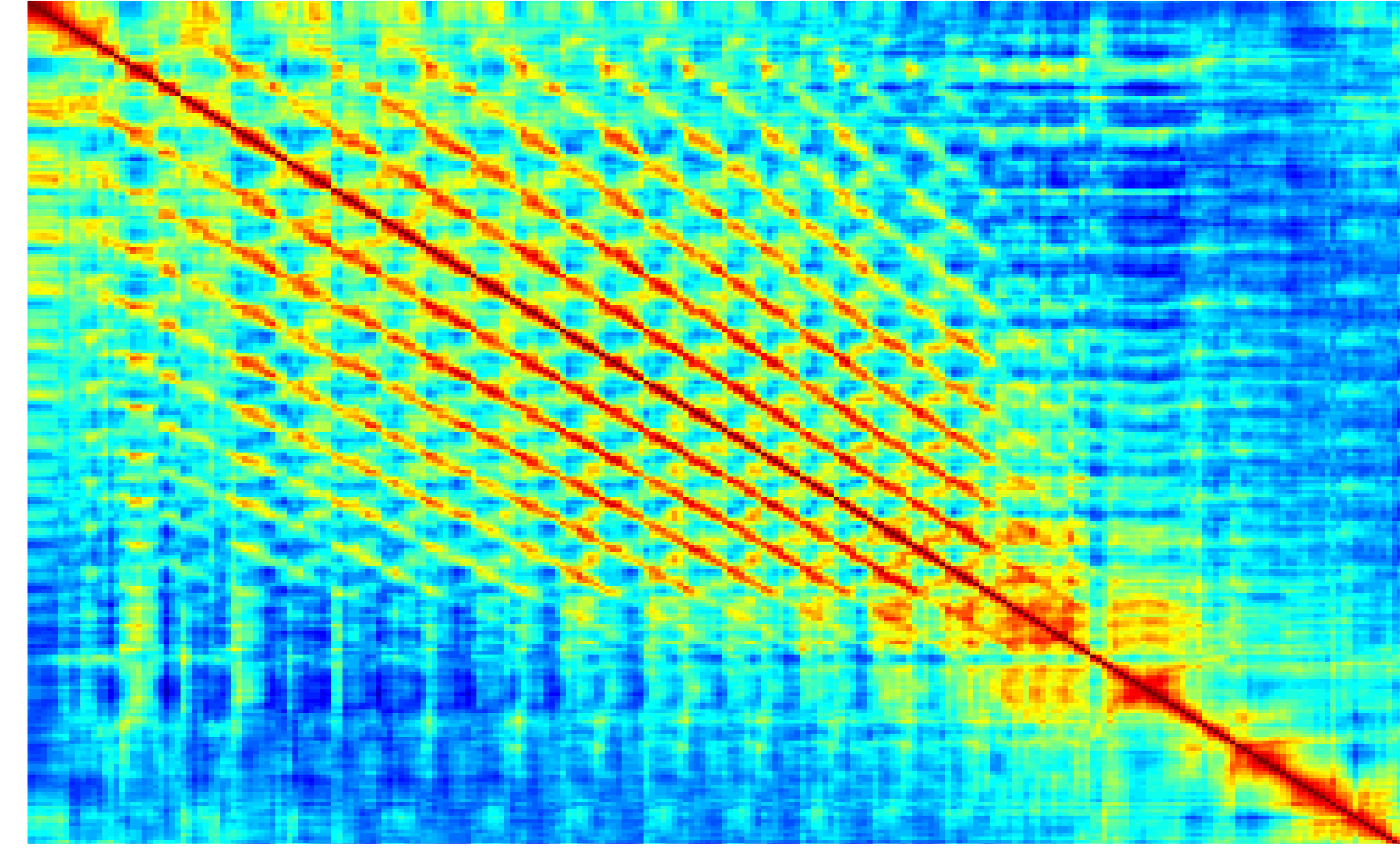}\\
    (a) & (b)  & (c)  & (d) \\
    \end{tabular}
    \caption{(a) Average ROC AUC for posture retrieval in the Olympic Sports dataset.
Similarities learned 
    by (b) 1-sample CNN, (c) using NN-CNN, and (d) for the proposed approach.
The plots show a magnified crop of the full similarity matrix.
Note the more detailed fine structure in (d).}
    \label{fig:roc_curves_selected}
\end{figure*}

To show the effectiveness of the proposed method we give empirical proof by training CNNs following both 1-sample CNN and NN-CNN training protocols.
Fig. \ref{fig:roc_curves_selected}(a) shows the average ROC curve for posture retrieval in the Olympic Sports dataset \cite{olympic_sports} (refer to Sec. \ref{sec:olympic_metrics} for further details) for 1-sample CNN, NN-CNN and the proposed method, which clearly outperforms both exemplar based strategies.
In addition, Fig. \ref{fig:roc_curves_selected}(b-d) show an excerpt of the similarity matrix learned for each method.
It becomes evident that the proposed approach captures more detailed similarity structures, e.g., the diagonal structures correspond to repetitions of the same gait cycle within a long jump.

\subsection{Initialization}

In the previous section we have shown the shortcomings of exemplar-based training of CNNs.
The key obstacle is the discrepancy between the single positive sample used in exemplar learning and the large amounts of data needed to train deep CNNs.
Therefore, given a single exemplar $\d_i$ we attempt to find an initial number of related samples to enable the training of a CNN which further improves the similarities between exemplars.
To obtain this initial group of related samples we employ LDA whitened HOG \cite{hoglda}, 
which is a fundamental and computationally efficient approach to estimate similarities $s_{ij}$ 
between large numbers of samples. Moreover, since they constitute a view-based approach, HOG features are viewpoint and rotation variant,
which is therefore beneficial for pose estimation in 2D.
We define $s_{ij}=s(\phi(\d_i),\phi(\d_j)) = \phi(\d_i)^\top \phi(\d_j)$,
where $\phi(\d_i)$ is the whitened HOG descriptor of the exemplar and $\S = (s_{ij}) \in \mathbb{R}^{N \times N}$ is the resulting kernel matrix.
The nearest neighbor of the sample $i$ is the sample $j$ which maximizes $s_{ij}$.

As can be seen from Fig.~\ref{fig:spectrum}(b) most of these similarities are evidently unreliable and, thus, the majority of samples cannot be properly ranked w.r.t. their similarity to an exemplar $\d_i$.
However, the most similar and most dissimilar samples can be reliably identified as they are sticking out from the similarity distribution.
We can thus utilize these samples to find a small set of nearest neighbors to the exemplar and a set of samples that are dissimilar. 

\subsection{Compact Cliques}\label{sec:cliqueFormation}

Given an exemplar $\d_i$, assigning the same label to its nearest neighbors (positive group) and another label to its furthest neighbors (negative group) is not suitable for learning similarities.
The exemplars in these groups may be close to $\d_i$ (or distant for the negative group) but not to another due to lacking transitivity.
Furthermore, simple synthetic augmentation of either the positive or negative groups \cite{exemplarcnn} 
does not add transitivity relations to other exemplars.
As a result, to learn intra-class similarities we need to restrict the model to groups of samples which are compact and 
mutually similar to another (i.e. a clique), where all samples in the clique are worthy of having the same label assigned.

To build candidate cliques we apply complete-linkage clustering to merge a $\d_i$ with its local neighborhood so that all these samples are mutually similar.
Therefore, we start at each $\d_i$ and merge the sample with its local neighborhood, so that all merged samples are mutually similar.
Thus, cliques are compact, differ in size, and may be mutually overlapping.
To reduce redundancy, highly overlapping cliques are subsequently merged by clustering cliques using farthest-neighbor clustering.
This agglomerative grouping is terminated if the intra-clique similarity of a cluster is less than half that of its constituents. 

Let $K$ be the resulting number of compact cliques and $N$ the number of samples $\d_i$.
Then $\C \in\{0,1\}^{K\times N}$ is the resulting assignment matrix of samples to cliques.

\subsection{Selecting Mutually Consistent Cliques}
\label{sec:batchselection}

After generating a set of compact cliques we assign a unique surrogate (i.e. artificial) label to each clique.
Which means that all the samples belonging to the same clique get the same surrogate label.
However, since only the highest and lowest similarities are reliable, samples in different cliques are not necessarily dissimilar, 
even if they get assigned a different surrogate label (e.g. cliques can partially overlap). 
This issue implies that the surrogate labeling is not consistent since samples with different surrogate labels can be highly similar. 
Motivated by this observation and leveraging the fact that CNNs are trained on batches of samples, we strive to find batches of mutually distant cliques to compose our batches. 
Thus, all samples in a batch can be labeled consistently because they are either similar (same compact clique) or dissimilar (different, distant clique). 
Samples with unreliable similarity then end up in different batches and we train a CNN successively on these batches. 

To find a set of different batches of mutually distant cliques we now design an optimization problem that produces a set of consistent batches of cliques. 
Let $\X \in \{0,1\}^{B \times K}$ be an indicator matrix that assigns $K$ cliques to $B$ batches 
(row $\x_{b}$ of $\X$ indicates the cliques in batch $b$) and 
$\S' \in \mathbb{R}^{K \times K}$ be the similarity between cliques (computed as the average pairwise sample similarity).
We enforce cliques in the same batch to be dissimilar by minimizing $\tr{(\X\S'\X^\top)}$.
Essentially, we seek a selection of cliques that minimize the sum of pairwise similarities 
between cliques for each batch $b$, integrated over all batches.
To remove the penalty for selecting compact cliques (i.e. with high self-similarity) we subtract $\tr{(\X\diag{(\S')}\X^\top)}$,
which defines the sum of similarities of cliques to themselves, integrated over all batches.
Moreover, each batch should maximize sample coverage, 
i.e., the number of distinct samples in all cliques of a batch $\|\x_{b}\C\|_{p}^{p}$ should be maximal. 
Finally, the number of distinct points covered by all batches, $\|\mathbb{1}\X\C\|_{p}^{p}$, should be maximal, 
so that the different (potentially overlapping) batches together comprise as many samples as possible. 
We select $p=1/16$
so that our penalty function roughly approximates the non-linear step function. 
The objective of the optimization problem then becomes
%
\begin{alignat}{4}
& \min_{\X\in\{0,1\}^{B\times K}} && \tr{(\X\S'\X^\top)}-&&\tr{(\X\diag{(\S')}\X^\top)}-\lambda_{1}\sum_{b=1}^{B}\|\x_{b}\C\|_{p_{}}^{p}-&&\lambda_{2}\|\mathbb{1}\X\C\|_{p_{}}^{p}  \label{eq:OPd}\\
& \text{s.t. } && \X\mathbb{1}_{K}^\top = r\mathbb{1}_{B}^\top   \label{eq:OPd_const}
\end{alignat}
where $r$ is the desired number of cliques in one batch for CNN training.
The number of batches, $B$, can be set arbitrarily high to allow for as many rounds of SGD training as desired.
If it is too low, this can be easily spotted as only limited coverage of training data can be achieved in the last term of Eq.~\eqref{eq:OPd}.
Since $\X$ is discrete, the optimization problem~\eqref{eq:OPd} is not easier than the 
Quadratic Assignment Problem which is known to be $NP$-hard~\cite{QAP}.
To overcome this issue we relax the binary constraints and force instead the continuous solution 
to the boundaries of the feasible range by maximizing the additional term $\lambda_{3}\|\X-0.5\|_{F}^2$ using the Frobenius norm.



\begin{figure*}
    \centering
    \setlength{\tabcolsep}{-1pt}
    \begin{tabular}{c c c c c c c c}
    \setlength{\fboxsep}{0pt}\setlength{\fboxrule}{3pt}
    \fbox{\includegraphics[height=71px]{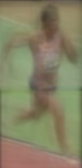}} &
    \setlength{\fboxsep}{0pt}\setlength{\fboxrule}{3pt}
    \fbox{\includegraphics[height=71px]{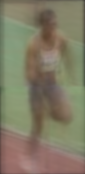}} &
    \setlength{\fboxsep}{0pt}\setlength{\fboxrule}{3pt}
    \fbox{\includegraphics[height=71px]{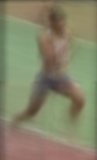}} &
    \setlength{\fboxsep}{0pt}\setlength{\fboxrule}{3pt}
    \fbox{\includegraphics[height=71px]{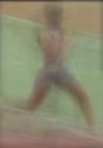}} &
    \setlength{\fboxsep}{0pt}\setlength{\fboxrule}{3pt}
    \fbox{\includegraphics[height=71px]{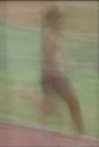}} &
    \setlength{\fboxsep}{0pt}\setlength{\fboxrule}{3pt}
    \fbox{\includegraphics[height=71px]{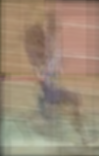}} &
    \setlength{\fboxsep}{0pt}\setlength{\fboxrule}{3pt}
    \fbox{\includegraphics[height=61px]{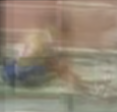}} &
    \setlength{\fboxsep}{0pt}\setlength{\fboxrule}{3pt}
    \fbox{\includegraphics[height=61px]{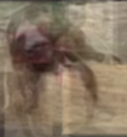}} \\
    \end{tabular}
    \caption{Visual example of a resulting batch of cliques for long jump category of Olympic Sports dataset.
Each clique contains at least 20 samples and is represented as their average.}
    \label{fig:batch}
\end{figure*}

We condition $\S'$ to be positive semi-definite by thresholding its eigenvectors and projecting onto the resulting base.
Since also $p<1$ the previous objective function is a difference of convex functions $u(\X)-v(\X)$, where 
\begin{alignat}{2}
u(\X) &= \tr{(\X\S'\X^\top)}-\lambda_{1}\sum_{b=1}^{B}\|\x_{b}\C\|_{p_{}}^{p}-\lambda_{2}\|\mathbb{1}\X\C\|_{p_{}}^{p}
\\
v(\X) &= \tr(\X\diag{(\S')}\X^\top) + \lambda_{3}\|\X-0.5\|_{F}^2  
\end{alignat}
It can be solved using the CCCP Algorithm \cite{CCCP}.
In each iteration of CCCP, the following convex optimization problem is solved,
\begin{alignat}{3}
    & \argmin_{\X\in [0,1]^{B\times K}} &&u(\X)-\vec{(\X)}^\top&&\vec{(\nabla v(\X^{t}))}, \label{eq:CCCP_iteration}\\
    & \text{s.t. } && \X \mathbb{1}_{K}^\top = r\mathbb{1}_{B}^\top \label{eq:CCCP_iteration_constr}
\end{alignat}
where $\nabla v(\X^{t}) = 2\X\odot(\mathbb{1}\diag{(\S')})+2\X-\mathbb{1}$ and $\odot$ denotes the Hadamard product.
We solve this constrained optimization problem by means of the interior-point method.
Fig. \ref{fig:batch} shows a visual example of a selected batch of cliques.

Let us now analyze the contribution of each term in Eq.~\eqref{eq:OPd}. 
We observed a significant drop in performance (by more than $30\%$) when we omitted any of the terms in Eq.~\eqref{eq:OPd} because of the following reasons:
\emph{(i)} omitting term $\left(\tr{(\X\S'\X^\top)} - \tr{(\X\diag{(\S')}\X^\top)}\right)$ will allow batches to have arbitrarily similar cliques.
Thus semantically very similar samples can occur in the same batch but with different labels;
\emph{(ii)} omitting term $\sum\limits_{b=1}^{B}\|\x_{b}\C\|_{p_{}}^{p}$ will allow a trivial solution -- each batch will degenerate to a single clique containing only a single sample;
\emph{(iii)} omitting term $\|\mathbb{1}\X\C\|_{p_{}}^{p}$ will yield $B$ identical batches, which contain the most dissimilar cliques.

\subsection{CNN Training} \label{sec:CNNTraining}

We successively train a CNN on the different batches $\x_b$ obtained by solving the minimization problem in Eq.~\eqref{eq:OPd}.
In each batch, classifying samples according to the clique they are in then serves as a pretext task for learning sample similarities.

One of the key properties of CNNs is the training using SGD and backpropagation \cite{backprop}.
The backpropagated gradient is estimated only over a subset (batch) of training samples, so it depends only on the subset of cliques in $\x_b$.
Following this observation, the clique categorization problem is effectively decoupled into a set of smaller sub-tasks (i.e. the individual batches of cliques).
During training, we randomly pick a batch $\x_b$ in each iteration and compute the stochastic gradient, using the loss $L(\W)$,

\begin{alignat}{2}
    L(\W) &\approx \frac{1}{M}\sum\limits_{j \in \x^{b}}f_{\W}(\d_j)&&+\lambda r(\W) \label{eq:loss} \\ 
    \V_{t+1} = \mu \V_{t} &- \alpha \nabla L(\W_t),\; &&\W_{t+1}=\W_{t}+ \V_{t+1} \; , \label{eq:grad} 
\end{alignat}

where $M$ is the SGD batch size, $\W_t$ denotes the CNN weights at iteration $t$, and $\V_t$ denotes the weight update of the previous iteration.
Parameters $\alpha$ and $\mu$ denote the learning rate and momentum, respectively.

We then compute similarities between exemplars by simply measuring correlation on the learned feature representation extracted from the CNN (see Sect. \ref{sec:olympic_metrics} for details).

\subsection{Local Temporal Pooling}\label{sec:TemporalContext}

The proposed approach as described so far models posture by exploiting sample (dis-)similarities of single images.
However, to learn the fine-grained similarities required to distinguish short-time actions, 
for instance, gait cycles of \textit{walking} vs. \textit{jogging}, not only posture matters but also how posture changes over short periods of time.
This means that not only similarities need to be exploited, but also temporal information has to be incorporated in the model in order to model fine-grained relationships.
Fortunately, a vast majority of the image data available for unsupervised learning contains this temporal information since it exists in the form of video sequences (e.g. YouTube videos),
which can be seen as sequences of exemplars $v_i~=~\{\d^{i}_{1}, \d^{i}_{2}, \dots, \d^{i}_{q}\}$ and $\d^{i}_{j}$ is the $j-$th exemplar (i.e. $j-$th frame) of the $i-$th video sequence.

\begin{table*}
    \scriptsize
    \centering
    \begin{tabular}{|c|c|c|c|c|}
    \hline
    HOG-LDA \cite{hoglda} & Ex-SVM \cite{exemplarsvm} & Ex-CNN \cite{exemplarcnn} & Alexnet \cite{alexnet} & 1-s CNN \\
    \hline
    0.62 & 0.72 & 0.64 & 0.65 & 0.67 \\
    \hline
    \end{tabular}
    \begin{tabular}{|c|c|c|c|c|c|c|c|c|}
    \hline
    NN-CNN & Doersch et. al \cite{ConvNetpretext1} & Shuffle\&Learn \cite{shuffleandlearn} & \textbf{Ours} & \textbf{Ours + LTP} \\
    \hline
    0.69 & 0.62 & 0.63 & \textbf{0.83} & \textbf{0.84} \\
    \hline
    \end{tabular}
    
    \caption{Avg. ROC AUC for each method on Olympic Sports dataset.}
    \label{tab:avg_auc}
\end{table*}

In this paper, we introduce an effective approach to incorporate temporal information in our model by performing a local average pooling of the exemplar similarities on the temporal dimension.
Given a pair of exemplars appearing in two different video sequences $(v_i, v_j)$, computing a simple global pooling over the entire sequence, 
as typically done for action classification \cite{globalpooling}, will result in losing fine-grained similarity structures over sub-sequences.
In addition, modelling temporal context with complex recurrent architectures like LSTMs \cite{lstm} has proven useful for action classification.
However, the temporal context that LSTMs encode cannot be learned for each exemplar, 
given a large number of exemplars available for unsupervised learning (e.g. the number of exemplars used in our experiments is in the order of $10^5$).

To overcome these issues, we locally pool the similarities in a small temporal neighborhood 
(i.e. a short sub-sequence) of $p$ frames around each exemplar.
Formally, let $s = \phi'(\d^{i}_{k})^\top \phi'(\d^{j}_{l})$ be the similarity between two exemplars,
where $\phi'$ is the feature representation learned by the CNN.
Then, the similarity obtained by employing temporal average pooling is defined as:

\begin{equation}
\centering
s' = \frac{1}{2p+1}\sum_{n \in \{-p, +p\}}\phi'(\d^{i}_{k+n})^\top \phi'(\d^{j}_{l+n})
\end{equation}

This method of modeling temporal context is fast and effective, giving us a boost in performance (cf. Sect. \ref{sec:olympic_metrics}, \ref{sec:ucf}) when temporal information is available in the dataset.

\subsection{Multiple Instance Learning of Similarities}

After a training round over all batches and performing locally temporal pooling we impute the similarities using the representation learned by the CNN.
This is motivated by the fact that once the training process converges, the similarities that are learned are more reliable than the ones used for initialization, 
and thus, enable the grouping algorithm from Sect. \ref{sec:cliqueFormation} to find larger cliques of mutually related samples. 

Since the number of unreliable similarities decreases after training the CNN, more samples can be comprised in a training batch and overall fewer batches already cover the same fraction of data as before training the CNN.
Therefore, we alternately train the CNN, perform locally temporal pooling on the resulting similarities and recompute cliques and batches using the similarities inferred in the previous step.
This alternating imputation of similarities and training of the CNN follows the idea of multiple-instance learning and has shown to converge in less than four iterations.

\begin{figure*}
\centering
  \begin{minipage}[c]{0.31\textwidth}
  \centering
    \includegraphics[width=\textwidth]{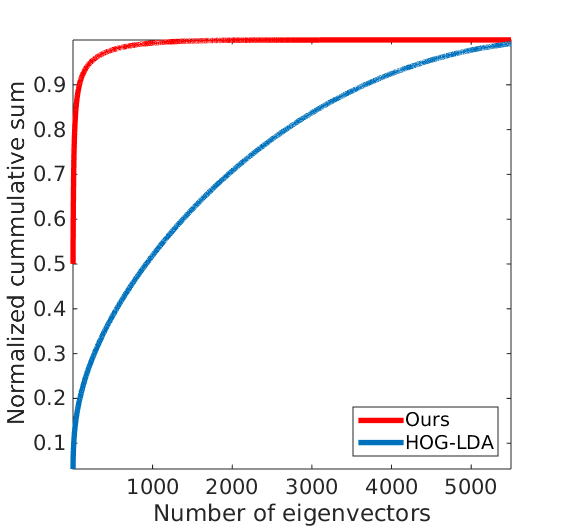}\\
    (a)
  \end{minipage}
  \hspace{-0.0cm}
    \begin{minipage}[c]{0.51\textwidth}
     \centering
    \includegraphics[width=\textwidth]{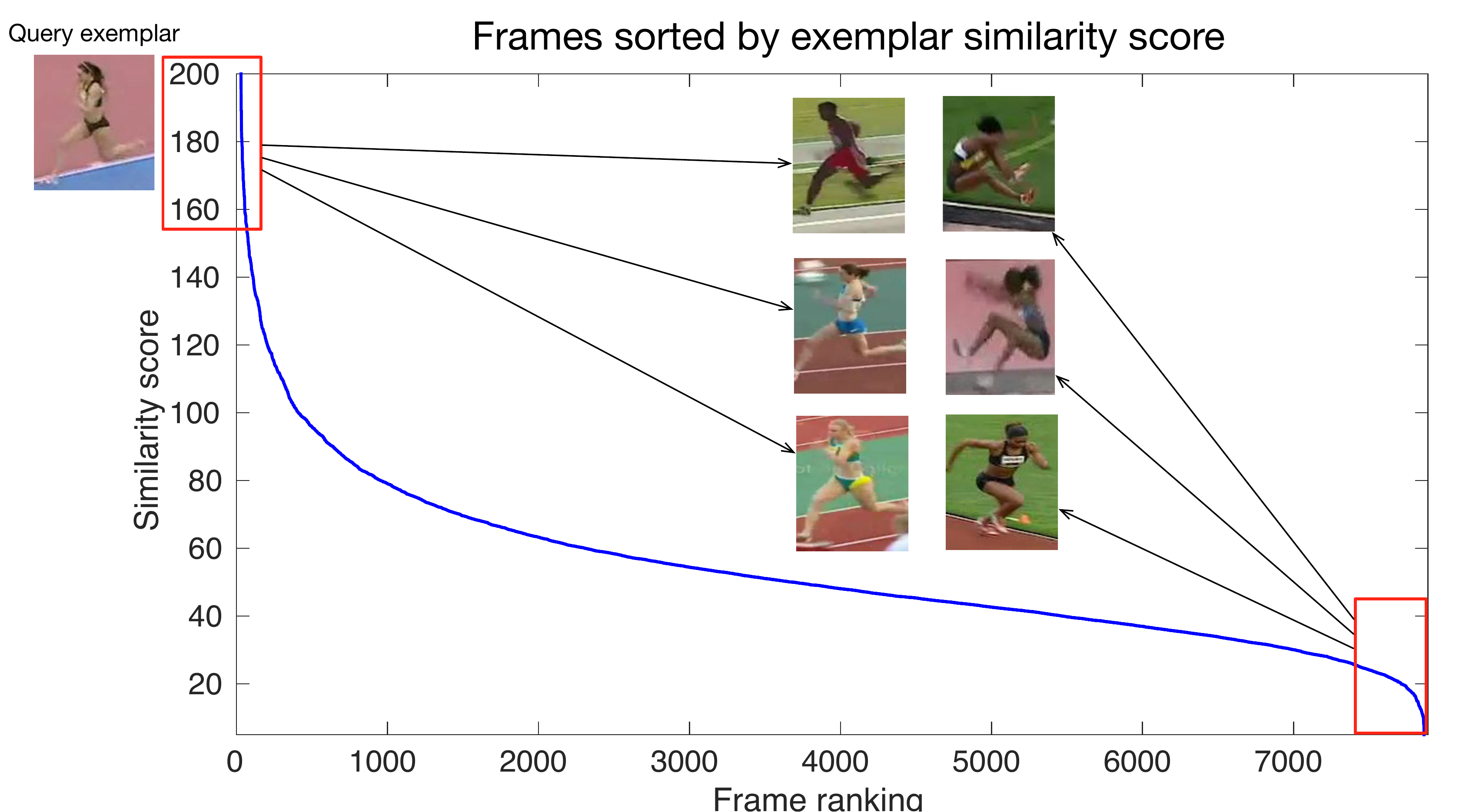}\\
    (b)
  \end{minipage}
\caption{(a) Cumulative distribution of the spectrum of the similarity matrices obtained by our method and the HOG-LDA initialization.
(b) Sorted similarities with respect to one exemplar, where only similarities at the ends of the distribution can be trusted.
} \label{fig:spectrum}
\end{figure*}

To evaluate the improvement of the similarities Fig. \ref{fig:spectrum} analyzes the eigenvalue spectrum of $\S$ on the Olympic Sports dataset, see Sect. \ref{sec:olympic_metrics}.
The plot shows the normalized cumulative sum of the eigenvalues as a function of the number of eigenvectors.
Compared to the similarities used for initialization, transitivity relations are learned and the approach can generalize from an exemplar to more related samples.
Therefore, the similarity matrix becomes more structured (cf. Fig.\ \ref{fig:roc_curves_selected}) and random noisy relations disappear.
As a consequence, it can be represented using very few basis vectors.

\section{Experimental Evaluation}\label{sec:experiments}

To compare our exemplar-based approach for unsupervised similarity learning with previous works we perform both quantitative and qualitative analysis.
We conduct experiments on unsupervised fine-grained posture retrieval on $3$ different Sports datasets: Olympic Sports \cite{olympic_sports}, UCF Sports \cite{ucf_sports1} and Leeds Sports Pose \cite{lsp}.
Furthermore, to demonstrate the capabilities of our model in the semi-supervised scenario we also tackled pose estimation on Leeds Sports \cite{lsp} and MPII Pose Dataset \cite{mpii}.
Finally, provided the wide applicability of the proposed approach we also undertake the unsupervised setup of object classification (PASCAL VOC 2007 dataset \cite{voc}). 

\subsection{Olympic Sports Dataset: Posture Analysis}\label{sec:olympic_metrics}
The Olympic Sports dataset \cite{olympic_sports} consists of video sequences of athletes practicing $16$ different sports.
The dataset contains an overall number of $113\,516$ frames, covering a rich set of human postures.
As we aim to evaluate fine-grained pose similarity, we had independent annotators manually label $20$ positive (similar) and negative (dissimilar) frames for $1033$ query exemplars.
We want to emphasize that these annotations are solely used for testing since our approach is unsupervised and does not utilize any labels during training.

\begin{figure}[!t]
    \centering
    \setlength{\tabcolsep}{-0.0pt}
    \begin{tabular}{c c c c}
    Query & Ours & Alexnet \cite{alexnet} & HOG-LDA \cite{hoglda}\\
    \includegraphics[width=0.25\columnwidth]{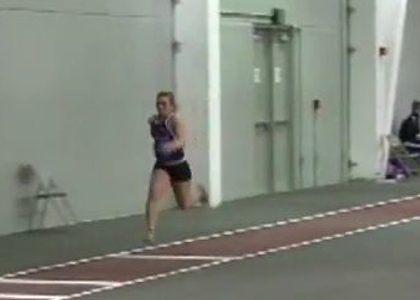} & \includegraphics[width=0.25\columnwidth]{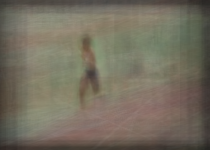} & \includegraphics[width=0.25\columnwidth]{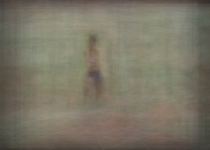} & 
    \includegraphics[width=0.25\columnwidth]{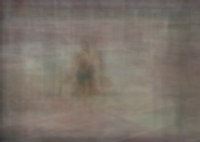}
    \end{tabular}
    \caption{Averaging of the 50 nearest neighbors for a given query frame using similarities obtained by our approach, Alexnet\cite{alexnet} and HOG-LDA \cite{hoglda}.}
    \label{fig:blending}
\end{figure}

We consider the following baselines for comparison with the proposed approach: HOG-LDA \cite{hoglda}, Exemplar-SVMs \cite{exemplarsvm}, 
Imagenet pretrained Alexnet \cite{alexnet}, 1-sample CNN and NN-CNN models (in a very similar spirit to \cite{clusteredsvm}),
Exemplar-CNN \cite{exemplarcnn}, the two-stream approach of Doersch et. al \cite{ConvNetpretext1}, and Shuffle\&Learn \cite{shuffleandlearn}.
To compute person bounding boxes we use the approach of \cite{dpm} as it shows 
reasonable performance in object and person detection.
\emph{(i)} The evaluation must explore the benefit of the unsupervised selecting of batches of cliques for deep learning of exemplars using standard CNN architectures.
For that reason, we incarnate our approach by adopting the widely used architecture of Krizhevsky et al.\ \cite{alexnet}.
To build batches for training the neural network we solve the optimization problem in Eq.~\eqref{eq:OPd} with $B=100$, $K=100$, and $r=20$ and fine-tune the model for $10^5$ iterations.
For the temporal average pooling we took a temporal neighborhood of $3$ frames around each exemplar.
After that we measure similarities using features extracted from layer fc7 in the \textit{caffe} implementation of \cite{alexnet}. 
\emph{(ii)} Exemplar-CNN is trained using the best performing parameters reported in \cite{exemplarcnn} and the 64c5-128c5-256c5-512f architecture.
Then we extract fc4 features and compute 4-quadrant max pooling.
\emph{(iii)} Exemplar-SVM is trained on the query exemplars using the HOG descriptor.
Hard negative mining is run on all the samples from all sports categories except the one which the exemplar belongs to.
We find an optimal number of negative mining rounds (less than three) using cross-validation and set the class weights of the linear SVM as $C_1=0.5$ and $C_2=0.01$.
\emph{(iv)} We compute LDA whitened HOG using approach from \cite{hoglda}.
\emph{(v)} The 1-sample CNN is trained by defining a separate class for each exemplar sample plus one class containing all other samples.
\emph{(vi)} In a similar fashion, the NN-CNN is trained using the exemplar plus $10$ nearest neighbors obtained using the whitened HOG similarities.
Both CNNs were implemented using the model of \cite{alexnet} and fine-tuning it for $10^5$ iterations. 
We employ AdaGrad  \cite{adagrad} solver with a batch size of $128$, learning rate of $0.001$ and smoothing term of $0.0001$. 
Each image in the training set was augmented with $10$ transformed versions by performing random translation, scaling, rotation and color transformation, to improve invariance with respect to these.


In Tab.~\ref{tab:avg_auc} we report the average AUC for each method over all categories of the Olympic Sports dataset.
 More specifically, the experiments witness that the 1-sample CNN fails to model the positive distribution, due to the high imbalance between positives and negatives and the resulting biased gradient.
In contrast, extra nearest neighbors to the exemplar (NN-CNN) yield a better model of the intra-class variability of the exemplar leading to a $2\%$ performance boost over the 1-sample CNN.
However, NN-CNN also sees a large set of negatives, which are partially similar and dissimilar.
Due to lack of structure in the negative set, NN-CNN fails to thoroughly capture the fine-grained similarities in the negative samples.
To avoid this issue we compute sets of mutually distant compact cliques resulting in a performance increase of $14\%$ over NN-CNN.
In addition, using locally temporal pooling (LTP) on the exemplar similarities with a neighborhood radius $p=3$ yields a further improvement of $1\%$.

\begin{figure*}[t]
    \centering
    \newcommand{\dpimgw}{0.1\textwidth}
    \setlength{\tabcolsep}{-0.0pt}
    \begin{tabular}{c||cccccccc}
    \textbf{Query} & & & & \textbf{NNs} & & & \\

    \includegraphics[width=\dpimgw]{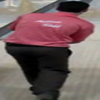} & 
    \includegraphics[width=\dpimgw]{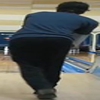} & 
    \includegraphics[width=\dpimgw]{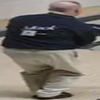} & 
    \includegraphics[width=\dpimgw]{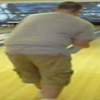} & 
    \includegraphics[width=\dpimgw]{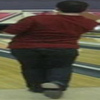} & 
    \includegraphics[width=\dpimgw]{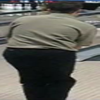} & 
    \includegraphics[width=\dpimgw]{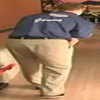} & 
    \includegraphics[width=\dpimgw]{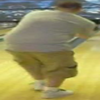} & 
    \includegraphics[width=\dpimgw]{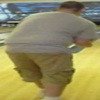} \\
    
    \includegraphics[width=\dpimgw]{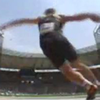} & 
    \includegraphics[width=\dpimgw]{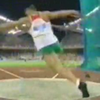} & 
    \includegraphics[width=\dpimgw]{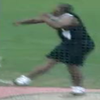} & 
    \includegraphics[width=\dpimgw]{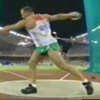} & 
    \includegraphics[width=\dpimgw]{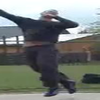} & 
    \includegraphics[width=\dpimgw]{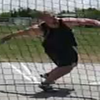} & 
    \includegraphics[width=\dpimgw]{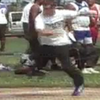} & 
    \includegraphics[width=\dpimgw]{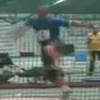} & 
    \includegraphics[width=\dpimgw]{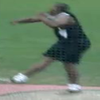} \\

    \includegraphics[width=\dpimgw]{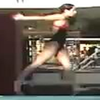} &
    \includegraphics[width=\dpimgw]{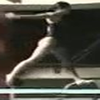} & 
    \includegraphics[width=\dpimgw]{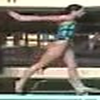} & 
    \includegraphics[width=\dpimgw]{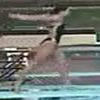} & 
    \includegraphics[width=\dpimgw]{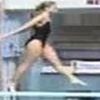} & 
    \includegraphics[width=\dpimgw]{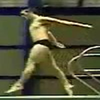} & 
    \includegraphics[width=\dpimgw]{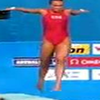} & 
    \includegraphics[width=\dpimgw]{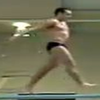} &
    \includegraphics[width=\dpimgw]{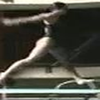} \\
    
    \includegraphics[width=\dpimgw]{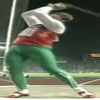} &
    \includegraphics[width=\dpimgw]{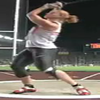} &
    \includegraphics[width=\dpimgw]{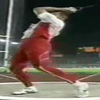} &
    \includegraphics[width=\dpimgw]{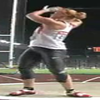} &
    \includegraphics[width=\dpimgw]{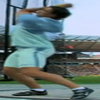} &
    \includegraphics[width=\dpimgw]{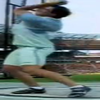} &
    \includegraphics[width=\dpimgw]{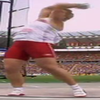} &
    \includegraphics[width=\dpimgw]{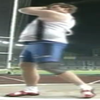} &
    \includegraphics[width=\dpimgw]{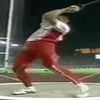} \\
    
    \includegraphics[width=\dpimgw]{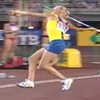} &
    \includegraphics[width=\dpimgw]{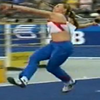} &
    \includegraphics[width=\dpimgw]{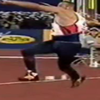} &
    \includegraphics[width=\dpimgw]{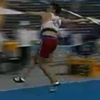} &
    \includegraphics[width=\dpimgw]{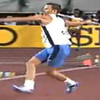} &
    \includegraphics[width=\dpimgw]{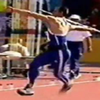} &
    \includegraphics[width=\dpimgw]{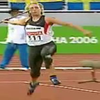} &
    \includegraphics[width=\dpimgw]{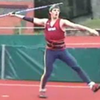} &
    \includegraphics[width=\dpimgw]{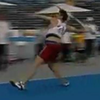} \\
    
    \includegraphics[width=\dpimgw]{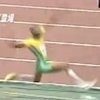} &
    \includegraphics[width=\dpimgw]{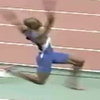} &
    \includegraphics[width=\dpimgw]{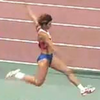} &
    \includegraphics[width=\dpimgw]{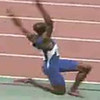} &
    \includegraphics[width=\dpimgw]{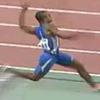} &
    \includegraphics[width=\dpimgw]{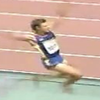} &
    \includegraphics[width=\dpimgw]{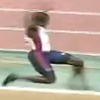} &
    \includegraphics[width=\dpimgw]{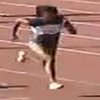} &
    \includegraphics[width=\dpimgw]{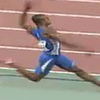} \\
    
    \includegraphics[width=\dpimgw]{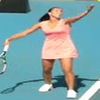} &
    \includegraphics[width=\dpimgw]{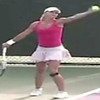} &
    \includegraphics[width=\dpimgw]{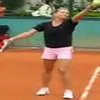} &
    \includegraphics[width=\dpimgw]{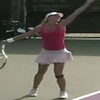} &
    \includegraphics[width=\dpimgw]{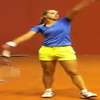} &
    \includegraphics[width=\dpimgw]{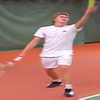} &
    \includegraphics[width=\dpimgw]{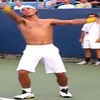} &
    \includegraphics[width=\dpimgw]{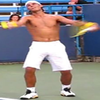} &
    \includegraphics[width=\dpimgw]{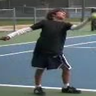} \\

    \end{tabular}
    \caption{Nearest neighbors retrieved by the proposed approach for representative query images of the Olympic Sports dataset.}
    \label{fig:nns_os}
\end{figure*}

Qualitatively, Fig.~\ref{fig:roc_curves_selected} renders the similarity matrices obtained by different approaches for a video sequence of the \textit{long jump} category.
In these matrices, the parallel diagonals indicate the gait cycle of a person running before leaping into the sandpit.
We can see how the method proposed in this paper clearly highlights these gait cycles while filtering noisy similarity relationships.
In addition, to visually assess the similarities we average the top $50$ nearest neighbors for a randomly chosen exemplar frame in the Olympic Sports dataset.
Fig.~\ref{fig:blending} shows how the neighbors obtained by our approach depict a sharper average posture since they come from compact cliques of mutually similar exemplar frames.
Therefore frames are more similar to the original and more details of the posture are retained than in case of the other methods.
Finally, in Fig.~\ref{fig:nns_os} we show nearest neighbors for few representative query images of the dataset.

\subsection{UCF Sports Dataset: Transferring Posture Representations}\label{sec:ucf}
\begin{table}[!t]
    \scriptsize
    \centering
    \begin{tabular}{|c|c|c|c|}
    \hline
    HOG-LDA \cite{hoglda} & Ex-SVM \cite{exemplarsvm} & Ex-CNN \cite{exemplarcnn} & Alexnet \cite{alexnet} \\ \hline
    0.67 & 0.71 & 0.68 & 0.68\\\hline
    \hline
    
    1-s CNN & NN-CNN & \textbf{Ours} & \textbf{Ours + LTP}\\ \hline
    0.59 & 0.66  & \textbf{0.78} & \textbf{0.79}\\ \hline
    \end{tabular}
    \caption{Avg. ROC AUC for each method on UCF Sports dataset.}
    \label{tab:avg_auc_ucf}
\end{table}

The UCF Sports dataset \cite{ucf_sports1} contains a set of actions from various sports.
Originally, this dataset consists of $12$ categories.
We disregarded the categories in which the posture does not change (e.g. Horse Riding) keeping $7$: diving side, golf swing side, kicking (kicking-front and kicking-side were merged together), weight lifting, run side, swing bench, swing side angle.
Having a total of $5148$ frames over all categories, which fails to fulfill with data volumes required to train deep CNN models, as the one proposed in this paper.
In this scenarios, where little training data is available, transfer learning has been proved to be a useful approach.

\begin{figure*}[!t]
    \centering
    \setlength{\tabcolsep}{-0.0pt}
    \begin{tabular}{c}
    \includegraphics[width=0.909\textwidth]{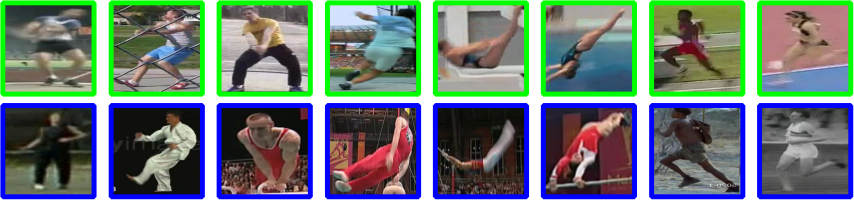} 
    \end{tabular}
    \caption{Based on the similarity structure learned by our model on Olympic Sports, postures are matched between Olympic (top row) and UCF (bottom row) Sports dataset.
At the bottom is the most similar frame to the one at the top.}
    \label{fig:ufc_olympic_sports_mathed_categories}
\end{figure*}

Therefore, we leverage the bigger Olympic Sports dataset and transfer the models learned on Olympic Sports categories using them solely for computing similarities of on the data of the UCF Sports dataset.
We visually matched $4$ categories of Olympic Sports to UCF Sports and transfer the learned models: hammer-throw and kicking, hammer-throw and swing-bench, diving-springboard-3m and swing-side-angle, long-jump and run-side.
A visual example of the matching postures between UCF Sports and Olympic Sports dataset is shown in Fig. \ref{fig:ufc_olympic_sports_mathed_categories}.

Analogously to the Olympic Sports dataset, independent annotators manually labeled $20$ positive (similar) and negative (dissimilar) frames for around $150$ exemplars in the above selected $4$ categories.
These annotations are solely used for testing since we do not train on UCF Sports dataset at all.

\begin{figure*}[t]
    \centering
    \setlength{\tabcolsep}{-0.0pt}
    \begin{tabular}{c c @{\hskip 1px} c c}
    \includegraphics[height=94px]{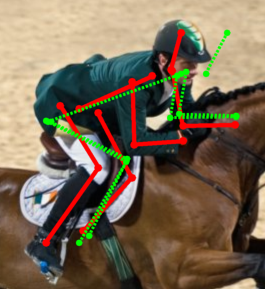} & 
    \includegraphics[height=94px]{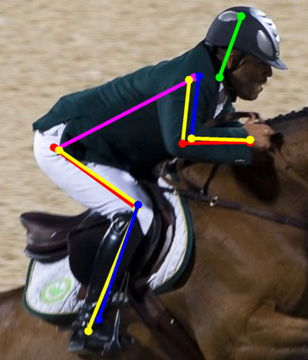} & 
    \includegraphics[height=94px]{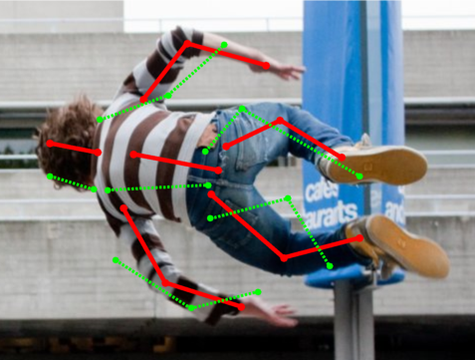} & 
    \includegraphics[height=94px]{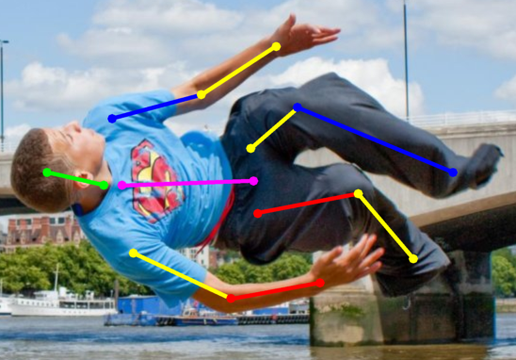}\\
        (a)  & (b) & (c) & (d) \\
    \end{tabular}
    \caption{Pose prediction results. (a) and (c) are test images with the superimposed ground truth skeleton depicted in red and the predicted skeleton in green.
(b) and (d) are corresponding nearest neighbors, which were used to transfer pose.}
    \label{fig:lsp_prediction}
\end{figure*}

\begin{figure*}[!t]
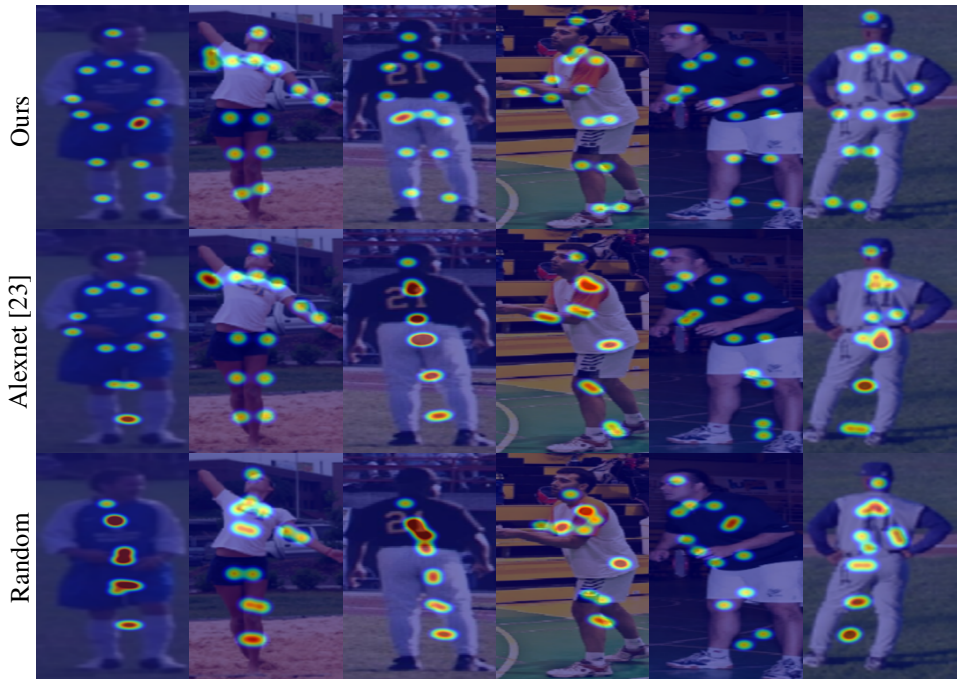

    \centering
    \newcommand{\dpimgw}{0.11}
    \newcommand{\dpimgh}{0.162}
    \newcommand{\inclfig}[1]{\includegraphics[width=\dpimgw\textwidth, height=\dpimgh\textwidth]{#1}}
    \setlength{\tabcolsep}{1.0pt}
    \renewcommand{\arraystretch}{0.0}
    
    \begin{tabular}{c c}
    
    \rotatebox{90}{\kern1.1cm Ours} & 
    \foreach \n in {41, 425, 505, 511, 553, 556} {\inclfig{img/deeppose/lsp_\n_deeppose_ours_7.png}} \\
    \rotatebox{90}{\kern0.6cm Alexnet \cite{alexnet}} & 
    \foreach \n in {41, 425, 505, 511, 553, 556} {\inclfig{img/deeppose/lsp_\n_deeppose_alexnet_7.png}} \\
    \rotatebox{90}{\kern1cm Random} & 
    \foreach \n in {41, 425, 505, 511, 553, 556} {\inclfig{img/deeppose/lsp_\n_deeppose_scratch_7.png}} \\
    \vspace{5pt} & \\
    
    
    \end{tabular}
    \caption{Heatmaps obtained by DeepPose (stg-1) \cite{deeppose} trained on LSP using different models as initialization.}
    \label{fig:lsp_deeppose}
\end{figure*}

\begin{table*}[!t]
    \scriptsize
    \centering
    \begin{tabular}{|c|c|c|c|c|c|c|c|}
    \hline
    Method  &Torso &Upper legs &Lower legs &Upper arms &Lower arms &Head &Total \\
    \hline
    Ours  & 80.1 & 50.1 & 45.7 & 27.2 & 12.6 &  45.5 &  43.5 \\
    \hline
    HOG-LDA\cite{hoglda}& 73.7 & 41.8 & 39.2 & 23.2 & 10.3 & 42.2& 38.4 \\
    \hline
    Shuffle\&Learn \cite{shuffleandlearn}  & 60.4  & 33.2 & 28.9 & 16.8 & 7.1 &  33.8 &  30.0\\
    \hline
    Alexnet\cite{alexnet}& 76.9 & 47.8 & 41.8 & 26.7 & 11.2 & 42.4 & 41.1 \\
    \hline
    Ground Truth & \textbf{93.7 } & 78.8 & 74.9 & 58.7 & 36.4 & 72.4 & 69.2\\
    \hline
    \hline
    \tiny{Pose Machines} \cite{posemachines} & 93.1 & \textbf{83.6} & \textbf{76.8} & \textbf{68.1 }& \textbf{42.2} & \textbf{85.4} & \textbf{72.0} \\
    \hline
    \end{tabular}
    \caption{PCP measure for each method on Leeds Sports dataset, using the retrieval based estimation for joint positions.}
    \label{tab:results_lsp}
\end{table*}

\begin{figure*}[!t]
    \centering
    \def\arraystretch{1.0}
    \newcommand{\dpimgw}{0.112\textwidth}
    \setlength{\tabcolsep}{-0.0pt}
    \begin{tabular}{c||ccccccc}
    \textbf{Query} & & & & \textbf{NNs} & & & \\
    \includegraphics[width=\dpimgw]{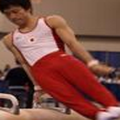} &
    \includegraphics[width=\dpimgw]{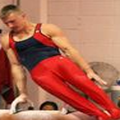} &
    \includegraphics[width=\dpimgw]{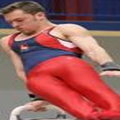} &
    \includegraphics[width=\dpimgw]{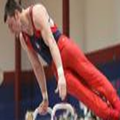} &
    \includegraphics[width=\dpimgw]{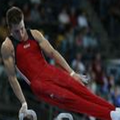} &
    \includegraphics[width=\dpimgw]{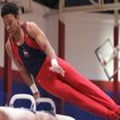} &
    \includegraphics[width=\dpimgw]{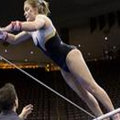} &
    \includegraphics[width=\dpimgw]{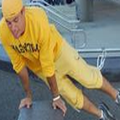} \\

    \includegraphics[width=\dpimgw]{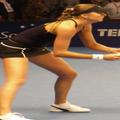} &
    \includegraphics[width=\dpimgw]{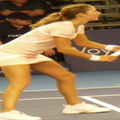} &
    \includegraphics[width=\dpimgw]{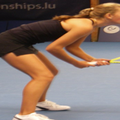} &
    \includegraphics[width=\dpimgw]{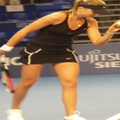} &
    \includegraphics[width=\dpimgw]{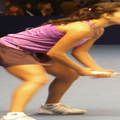} &
    \includegraphics[width=\dpimgw]{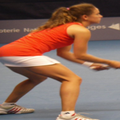} &
    \includegraphics[width=\dpimgw]{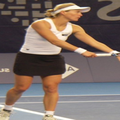} &
    \includegraphics[width=\dpimgw]{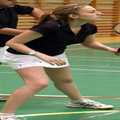} \\
    
    \includegraphics[width=\dpimgw]{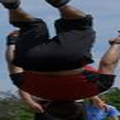} &
    \includegraphics[width=\dpimgw]{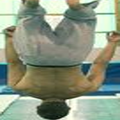} &
    \includegraphics[width=\dpimgw]{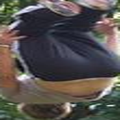} &
    \includegraphics[width=\dpimgw]{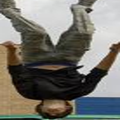} &
    \includegraphics[width=\dpimgw]{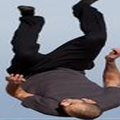} &
    \includegraphics[width=\dpimgw]{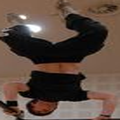} &
    \includegraphics[width=\dpimgw]{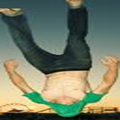} &
    \includegraphics[width=\dpimgw]{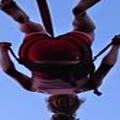} \\

    \includegraphics[width=\dpimgw]{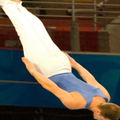} &
    \includegraphics[width=\dpimgw]{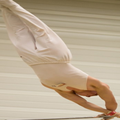} &
    \includegraphics[width=\dpimgw]{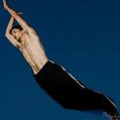} &
    \includegraphics[width=\dpimgw]{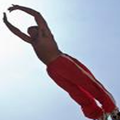} &
    \includegraphics[width=\dpimgw]{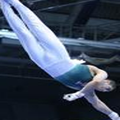} &
    \includegraphics[width=\dpimgw]{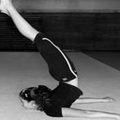} &
    \includegraphics[width=\dpimgw]{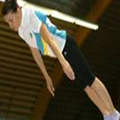} &
    \includegraphics[width=\dpimgw]{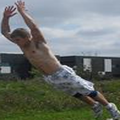} \\
    
    \includegraphics[width=\dpimgw]{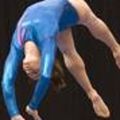} &
    \includegraphics[width=\dpimgw]{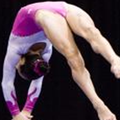} &
    \includegraphics[width=\dpimgw]{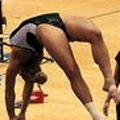} &
    \includegraphics[width=\dpimgw]{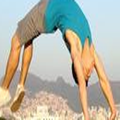} &
    \includegraphics[width=\dpimgw]{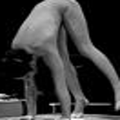} &
    \includegraphics[width=\dpimgw]{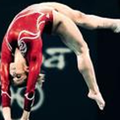} &
    \includegraphics[width=\dpimgw]{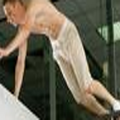} &
    \includegraphics[width=\dpimgw]{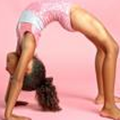} \\
    
    \includegraphics[width=\dpimgw]{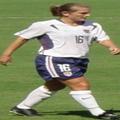} &
    \includegraphics[width=\dpimgw]{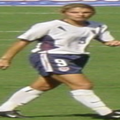} &
    \includegraphics[width=\dpimgw]{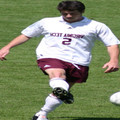} &
    \includegraphics[width=\dpimgw]{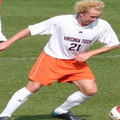} &
    \includegraphics[width=\dpimgw]{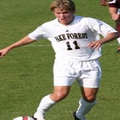} &
    \includegraphics[width=\dpimgw]{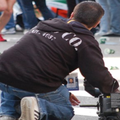} &
    \includegraphics[width=\dpimgw]{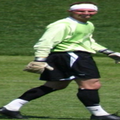} &
    \includegraphics[width=\dpimgw]{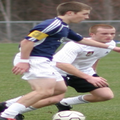} \\
    
    \includegraphics[width=\dpimgw]{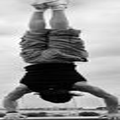} &
    \includegraphics[width=\dpimgw]{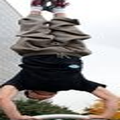} &
    \includegraphics[width=\dpimgw]{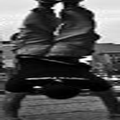} &
    \includegraphics[width=\dpimgw]{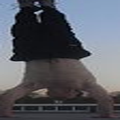} &
    \includegraphics[width=\dpimgw]{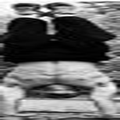} &
    \includegraphics[width=\dpimgw]{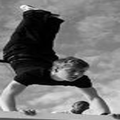} &
    \includegraphics[width=\dpimgw]{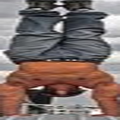} &
    \includegraphics[width=\dpimgw]{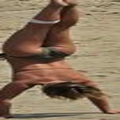} \\

    \end{tabular}
    \caption{Nearest neighbors retrieved by the proposed approach for representative query images of the Leeds Sports dataset.}
    \label{fig:nns_lsp}
\end{figure*}

\begin{table*}
    \scriptsize
    \centering
    \begin{tabular}{|c|c|c|c|c|c|c|c|}
    \hline
    Initialization  &Torso &Upper legs &Lower legs &Upper arms &Lower arms &Head &Total \\
    \hline
    \textbf{Ours}  & \textbf{93.9} & \textbf{71.2} & \textbf{55.0} & \textbf{44.5} & \textbf{21.6} & \textbf{63.2} &  \textbf{58.2}\\
    \hline
    Shuffle\&Learn \cite{shuffleandlearn} & 90.4 & 62.7 & 45.7 & 33.3 & 11.8 & 52.0 &  49.3\\
    \hline
    Random & 87.3 & 52.3 & 35.4 & 25.4 & 7.6 & 44.0 &  42.0\\
    \hline
    \hline
    Alexnet \cite{alexnet} & 92.8 & 68.1 & 53.0 & 39.8 & 17.5 & 62.8 &  55.7\\
    \hline
    \end{tabular}
    \caption{PCP measure for each method on Leeds Sports dataset using different models as initialization for training DeepPose \cite{deeppose}.}
    \label{tab:results_lsp_deeppose}
\end{table*}

\subsection{Leeds Sports Dataset: Pose Estimation}
\label{subsec:lsp}

We report the average ROC AUC for our approach, Exemplar-CNN \cite{exemplarcnn}, 1-sample CNN, NN-CNN models, Alexnet \cite{alexnet}, Exemplar-SVMs \cite{exemplarsvm}, and HOG-LDA \cite{hoglda}.
For each of the CNN-based approaches we simply transfer the learned representations from the matched categories of the Olympic Sports dataset, so no additional training is required.
The experimental settings are the ones described in Sect. \ref{sec:olympic_metrics}.

Tab.~\ref{tab:avg_auc_ucf} shows the average AUC for each of the compared methods on the $4$  categories of UCF Sports.
In particular, our approach attains a significant performance improvement of at least $7\%$ with respect to all compared methods.
Furthermore, when temporal information is incorporated in the model by pooling the similarities using locally temporal pooling we obtain a further improvement of $1\%$.
These results support the fact that the feature representation learned by our approach in Olympic Sports encodes a general notion of  posture, 
and therefore can be transferred without requiring any further learning to different categories of the UCF dataset.

The Leeds Sports Pose (LSP) Dataset \cite{lsp} is a well-known and widely used benchmark for pose estimation.
This dataset consists of $1000$ images for training which we combine with $4000$ images from its extended version, where each image is annotated with all $14$ joint locations from a person-centric viewpoint.
Finally, the test set consists of $1000$ images. 

\begin{figure*}[!t]
    \centering
    \def\arraystretch{1.0}
    \newcommand{\dpimgw}{0.1\textwidth}
    \setlength{\tabcolsep}{-0.0pt}
    \begin{tabular}{c||cccccccc}
    \textbf{Query} & & & & \textbf{NNs} & & & \\
    \includegraphics[width=\dpimgw]{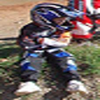} &  
    \includegraphics[width=\dpimgw]{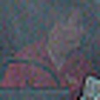} &  
    \includegraphics[width=\dpimgw]{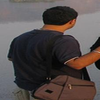} &  
    \includegraphics[width=\dpimgw]{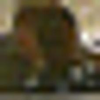} &  
    \includegraphics[width=\dpimgw]{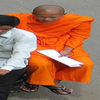} &  
    \includegraphics[width=\dpimgw]{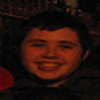} &  
    \includegraphics[width=\dpimgw]{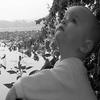} &  
    \includegraphics[width=\dpimgw]{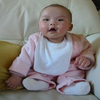} & 
    \includegraphics[width=\dpimgw]{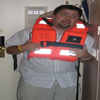} \\
    
    \includegraphics[width=\dpimgw]{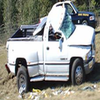} &  
    \includegraphics[width=\dpimgw]{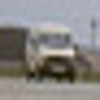} &  
    \includegraphics[width=\dpimgw]{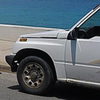} &  
    \includegraphics[width=\dpimgw]{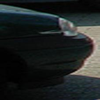} &  
    \includegraphics[width=\dpimgw]{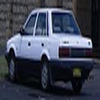} &  
    \includegraphics[width=\dpimgw]{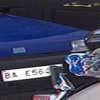} &  
    \includegraphics[width=\dpimgw]{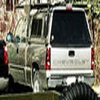} &  
    \includegraphics[width=\dpimgw]{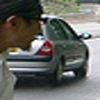} &
    \includegraphics[width=\dpimgw]{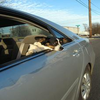} \\
    
    \includegraphics[width=\dpimgw]{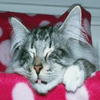} &  
    \includegraphics[width=\dpimgw]{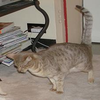} &  
    \includegraphics[width=\dpimgw]{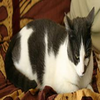} &  
    \includegraphics[width=\dpimgw]{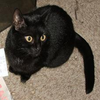} &  
    \includegraphics[width=\dpimgw]{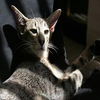} &  
    \includegraphics[width=\dpimgw]{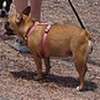} &  
    \includegraphics[width=\dpimgw]{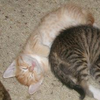} &  
    \includegraphics[width=\dpimgw]{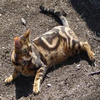} & 
    \includegraphics[width=\dpimgw]{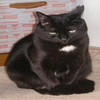} \\
    
    \includegraphics[width=\dpimgw]{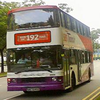} &  
    \includegraphics[width=\dpimgw]{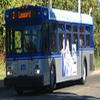} &  
    \includegraphics[width=\dpimgw]{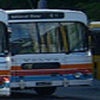} &  
    \includegraphics[width=\dpimgw]{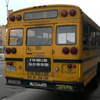} &  
    \includegraphics[width=\dpimgw]{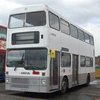} &  
    \includegraphics[width=\dpimgw]{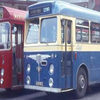} &  
    \includegraphics[width=\dpimgw]{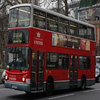} &  
    \includegraphics[width=\dpimgw]{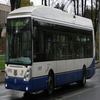} &
    \includegraphics[width=\dpimgw]{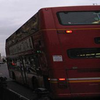} \\
    
    \includegraphics[width=\dpimgw]{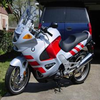} &  
    \includegraphics[width=\dpimgw]{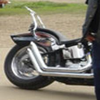} &  
    \includegraphics[width=\dpimgw]{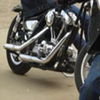} &  
    \includegraphics[width=\dpimgw]{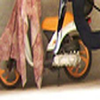} &  
    \includegraphics[width=\dpimgw]{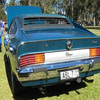} &  
    \includegraphics[width=\dpimgw]{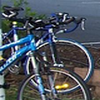} &  
    \includegraphics[width=\dpimgw]{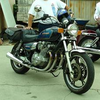} &  
    \includegraphics[width=\dpimgw]{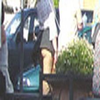} & 
    \includegraphics[width=\dpimgw]{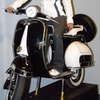} \\
    
    \includegraphics[width=\dpimgw]{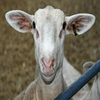} &  
    \includegraphics[width=\dpimgw]{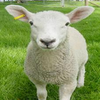} &  
    \includegraphics[width=\dpimgw]{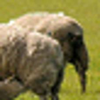} &  
    \includegraphics[width=\dpimgw]{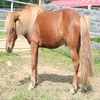} &  
    \includegraphics[width=\dpimgw]{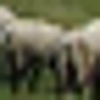} &  
    \includegraphics[width=\dpimgw]{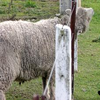} &  
    \includegraphics[width=\dpimgw]{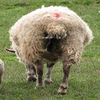} &  
    \includegraphics[width=\dpimgw]{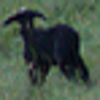} & 
    \includegraphics[width=\dpimgw]{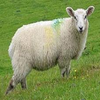} \\
    
    \includegraphics[width=\dpimgw]{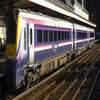} &  
    \includegraphics[width=\dpimgw]{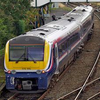} &  
    \includegraphics[width=\dpimgw]{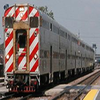} &  
    \includegraphics[width=\dpimgw]{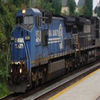} &  
    \includegraphics[width=\dpimgw]{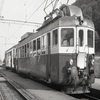} &  
    \includegraphics[width=\dpimgw]{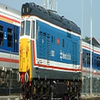} &  
    \includegraphics[width=\dpimgw]{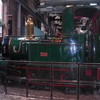} &  
    \includegraphics[width=\dpimgw]{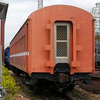} & 
    \includegraphics[width=\dpimgw]{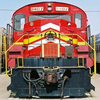} \\

    \end{tabular}
    \caption{Nearest neighbors retrieved by the proposed approach for representative query images of the VOC2007 dataset.}
    \label{fig:nns_voc}
\end{figure*}

We now evaluate the proposed approach on the problem of unsupervised pose estimation on LSP.
During training, we disregard all joint annotations from the training set and learn pose similarities.
During testing, these pose similarities yield by our approach are used to find frames similar in posture to a query frame.
The joint locations of the test image are then estimated by identifying its nearest neighbor from the training set and transferring its joint locations to the test image. 

For training our model we use the parameters described in Sect.~\ref{sec:olympic_metrics}.
The similarity between two images is measured as Pearson correlation on features extracted from layer fc6.
To evaluate the results we use the Percentage of Correct Parts (PCP) measure, which is the standard metric for benchmarking pose estimation methods.


For comparison with other methods, we follow the same testing protocol and retrieve similar postures using HOG-LDA \cite{hoglda}, and fc6 representations of Alexnet \cite{alexnet} and Shuffle\&Learn \cite{shuffleandlearn}.
In addition, we also report an upper bound on the performance that can be achieved by the nearest neighbor joint transfer, using ground-truth similarities to retrieve nearest neighbors.
Therefore, the nearest training pose for a test image is identified by minimizing the average Euclidean distance between their ground-truth pose annotation.
This is the best result one can achieve by finding the most similar pose, when not provided with a supervised parametric model (the performance gap to $100\%$ shows the degree of difference between training and test poses).
For completeness, we also compare with a fully supervised state-of-the-art approach for pose estimation \cite{posemachines}.
We use the same experimental settings described in Sect.~\ref{sec:olympic_metrics}.

The Percentage of Correct Parts @$0.5$ (PCP) for different approaches is reported in Tab.~\ref{tab:results_lsp}.
 In Tab.~\ref{tab:results_lsp} our approach improves the visual similarities learned using both Alexnet and HOG-LDA.
It is noteworthy that even though our approach for estimating the pose is \emph{fully unsupervised} it achieves a competitive performance when compared to the upper-bound of supervised ground truth similarities.
Qualitative results of nearest neighbors for several query frames are presented in Fig.~\ref{fig:nns_lsp}.


In addition, Fig.~\ref{fig:lsp_prediction} shows success (a) and failure (c) cases of our method.
In Fig.~\ref{fig:lsp_prediction}(a) we can see that the pose is correctly transferred from the nearest neighbor (b) from the training set, resulting in a PCP score of $0.6$ for that particular image.
Moreover, Fig.~\ref{fig:lsp_prediction}(c), (d) witness that our method learns the representation invariant to front-back flips (matching a person facing away from the camera to one facing the camera).
Since our approach learns pose similarity in an unsupervised manner, it becomes invariant to changes in appearance as long as the shape is similar, thus explaining this confusion.
Adding extra training data or directly incorporating face detection-based features could resolve this.

Furthermore, in addition to the fully unsupervised experiment, we evaluate the representation learned by the proposed approach on LSP, in a semi-supervised scenario, 
by using it as initialization for the supervised DeepPose \cite{deeppose} method. 
We train DeePose (stg-$1$) \cite{deeppose} with using different initializations: (a) random initialization, (b) Imagenet pre-trained AlexNet \cite{alexnet} (c) Shuffle\&Learn \cite{shuffleandlearn} and (d) our model trained on LSP dataset.
We then follow the training procedure described in \cite{deeppose}, where the train split includes $11000$ images (using the extended LSP data), and the test split includes $1000$ images.
 We use a batch size of $128$, learning rate of $5 \times 10^{-4}$ 
and optimize the CNN parameters using AdaGrad \cite{adagrad}.

\newcolumntype{C}[1]{>{\centering\let\newline\\\arraybackslash\hspace{0pt}}m{#1}}
\begin{table*}[!t]
    \scriptsize
    \centering
    \tabcolsep=0.04cm
    \tiny{
    \begin{tabular}{|c|C{0.76cm}|C{0.76cm}|*{8}{c|}|C{0.8cm}|}
    \hline
      & Head & Neck & LR Shoulder & LR Elbow & LR Wrist & LR Hip & LR Knee & LR Ankle & Thorax & Pelvis & Total\\
    \hline
    \textbf{Ours} & \textbf{89.5} & \textbf{93.7} & \textbf{85.9} & \textbf{71.6} & \textbf{56.3} & \textbf{82.7} & \textbf{72.4} & \textbf{67.3} & \textbf{93.8} & \textbf{88.3} & \textbf{80.2}\\
    \hline
    Shuffle\&Learn\cite{shuffleandlearn} & 75.8 & 86.3 & 75.0 & 59.2 & 42.2 & 73.3 & 63.1 & 51.7 & 87.1 & 79.5 & 69.3\\
    \hline
    Random & 79.5 & 87.1 & 71.6 & 52.1 & 34.6 & 64.1 & 58.3 & 51.2 & 85.5 & 70.1 & 65.4\\
    \hline
    \hline
    AlexNet\cite{alexnet} & 87.2 & 93.2 & 85.2 & 69.6 & 52.0 & 81.3 & 69.7 & 62.0 & 93.4 & 86.6 & 78.0\\
    \hline
    \end{tabular}
    }
    \caption{PCKh@$0.5$ measure for different limbs on MPII Pose benchmark dataset using different initializations for the DeepPose approach \cite{deeppose}.}
    \label{tab:results_mpii_deeppose}
\end{table*}

\begin{table*}[!t]
    \scriptsize
    \centering
    \begin{tabular}{|c |c| c | c | c|c|}
    \hline
     HOG-LDA & Wang et. al \cite{ConvNetpretext2} & Wang et. al \cite{ConvNetpretext2} + \textbf{Ours} & Alexnet \cite{alexnet} & RCNN\\ 
     \hline
         0.1180 &  0.4501  & 0.4812 & 0.6160 & 0.6825 \\
    \hline
    \end{tabular}
    \caption{Classification results for PASCAL VOC 2007}
    \label{tab:results_voc}
\end{table*}

Tab.~\ref{tab:results_lsp_deeppose} shows the PCP$@0.5$ score of DeepPose (stg-$1$) model trained using different methods as initialization.
Using our model to initialize DeepPose (stg-$1$) yields a performance boost of $2.5\%$ over Imagenet pretrained Alexnet\cite{alexnet} initialization.
Showing, as a result, that the representation learned by our model successfully encodes relevant pose information 
which is not only good for unsupervised pose retrieval but can further facilitate the training of supervised pose estimation methods.
Finally, in Fig. \ref{fig:lsp_deeppose} we show the predicted joint heatmaps obtained by DeepPose \cite{deeppose} when using the three different initialization models for several representative images of LSP dataset \cite{lsp}.

\subsection{MPII Dataset: Pose Estimation}

Next, to further assess the reliability and robustness of the pose representation learned by our model, we tackle the challenging MPII Pose dataset \cite{mpii}.
MPII Pose dataset \cite{mpii} is a state of the art benchmark for evaluation of articulated human pose estimation.
MPII Pose is a particularly challenging dataset because of the clutter, occlusion and number of persons appearing in images.
To evaluate our approach on MPII Pose we follow the semi-supervised training protocol used for LSP and compare the performance obtained by DeepPose (stg-$1$) \cite{deeppose}, 
when trained using as initialization each of the following models: Random initialization, Imagenet pre-trained AlexNet \cite{alexnet}, Shuffle\&Learn \cite{shuffleandlearn} and our approach trained on LSP in unsupervised manner (Sec.~\ref{subsec:lsp}). 
We use PCKh@$0.5$ on all the keypoints of the full body as evaluation metric which is the standard for MPII dataset.
Tab. \ref{tab:results_mpii_deeppose} reports the PCKh@$0.5$ obtained by the DeepPose (stg-1) models \cite{deeppose} with different initializations.
 In particular, when comparing our unsupervised initialization with a random initialization we obtain a $15\%$ performance boost, which indicates that our features encode a notion of pose that is robust to the clutter present in MPII dataset.
Furthermore, we obtain a $2.2\%$ improvement over Imagenet pre-trained AlexNet\cite{alexnet}.
The performance obtained on MPII Pose dataset corroborates that the representation learned by our method captures fine-grained posture details and successfully deals with clutter, 
occlusions and presence of multiple persons presented in this dataset.

\subsection{PASCAL VOC 2007: Object Classification}

Provided the wide applicability of our method, in addition to the experiments on pose estimation datasets in the previous sections we now evaluate the learning of similarities over object categories.
For this purpose, we classify object bounding boxes of the PASCAL VOC 2007 dataset \cite{voc}.
Instead of predicting the bounding box position and category, we assume that bounding boxes are given, 
provided recent outstanding results for object \cite{rcnn} and objectness \cite{objectness} detection, and focus directly on the object classification. 

To initialize our model we use the visual similarities of Wang et al. \cite{ConvNetpretext2} without applying any fine tuning on PASCAL and also compare against this approach.
Thus, neither Imagenet nor Pascal VOC labels are utilized during training or pre-training.
We then evaluate how our model performs in comparison with features obtained by HOG-LDA \cite{hoglda}, Wang et. al \cite{ConvNetpretext2},
Alexnet \cite{alexnet} pretrained on Imagenet and R-CNN \cite{rcnn} which is trained in a supervised manner on Pascal VOC.
For our method and HOG-LDA we use the same experimental settings as described in Sect. \ref{sec:olympic_metrics}.

At test time, we perform $k$-nearest neighbor classification for all methods.
The $k$ nearest neighbors are computed using similarities (Pearson correlation) based on the feature representation obtained for each method.
In Tab. \ref{tab:results_voc} we show the classification accuracies of all approaches for $k=5$ (for $k>5$ there was only insignificant performance improvement).
We can see how our approach improves upon the similarities of \cite{ConvNetpretext2} used as initialization to yield a performance gain of $3\%$ without requiring any supervision information or fine-tuning on PASCAL.
Finally, in Fig. \ref{fig:nns_voc} we show the retrieved nearest neighbors for few query samples of different categories of the Pascal VOC dataset \cite{voc}.


\section{Conclusion}\label{sec:conclusions}

In this manuscript we have proposed a technique for deep unsupervised learning of visual similarities between a large number of exemplars.
We analyze the shortcomings of exemplar learning on CNNs and address the single positive exemplar setup, the imbalance between exemplar and negatives, and inconsistent labels within SGD batches.
We address these key problems by optimizing a single cost function yielding SGD batches of compact, mutually dissimilar cliques of samples.
Each of these cliques then gets assigned a surrogate label, and the learning of visual similarities is then posed as a categorization task on individual batches.

In the experimental evaluation the proposed approach has shown competitive performance compared to the state-of-the-art, 
providing significantly finer similarity structure that is particularly crucial for detailed posture analysis.
Furthermore, the experimental evaluation in several pose datasets shows that the pose representation learned by our model in an unsupervised manner 
is transferable across pose datasets and can be used in conjunction with supervised parametric models for pose estimation to boost their performance.
Finally, the proposed approach also demonstrates competitive performance in general object classification problems.

\section*{Acknowledgements}
This research has been funded in part by the Heidelberg Academy of Sciences and a NVIDIA hardware grant.

\bibliographystyle{elsarticle-num}
\section*{References}
{\footnotesize
\bibliography{mybibfile}}

\vspace{0.1cm}
{\bf Artsiom Sanakoyeu} received a diploma in computer science from Belarusian State University, Minsk, Belarus in 2015.
He is currently a Ph.D. Candidate in Heidelberg Collaboratory for Image Processing at Heidelberg University.
His current research interests include computer vision and deep representation learning.

{\bf Miguel A. Bautista} earned his Ph.D. in Error-Correcting Representations for Multi-class problems at University of Barcelona in 2016.
His main research interest lay in the intersection of computer vision and machine learning.
He is particularly interested in high-level vision problems.

{\bf Björn Ommer}
received a diploma in computer science from the University of Bonn, Germany and Ph.D. in computer science from ETH Zurich, Switzerland in 2007. 
Since 2009 he is heading the computer vision group at Heidelberg University, 
where he serves as full professor for scientific computing in the Department of Mathematics and Computer Science. 
His research interests include computer vision, machine learning, and cognitive science.

\end{document}